\documentclass[11pt]{article}
\usepackage{CJKutf8}
 \usepackage{amsmath}
 \usepackage{bm}
\usepackage[final]{acl}
\usepackage{cellspace}
\usepackage{times}
\usepackage{latexsym}
\usepackage{enumitem}
\usepackage[normalem]{ulem}

\usepackage{newtxtext,newtxmath} 

\usepackage{microtype}  

\usepackage{inconsolata}

\usepackage{graphicx}

\usepackage{makecell}

\usepackage{tcolorbox}
\tcbuselibrary{breakable,skins} 
\usepackage{dblfloatfix}       
\usepackage{cuted} 

\usepackage{CJKutf8}
\usepackage{CJKutf8}
\usepackage{pdfrender}  

\newcommand{\mybold}[1]{%
  \textpdfrender{%
    TextRenderingMode=FillStroke,  
    LineWidth=0.2pt,              
  }{#1}%
}

\usepackage{CJKutf8}
\usepackage{xparse}

\title{Knowledge-Augmented Question Error Correction for Chinese Question Answer System with QuestionRAG}

\author{
Longpeng Qiu$^{1}$ \quad Ting Li$^{2}$ \quad Shuai Mao$^{2}$ \quad Nan Yang$^{2}$ \quad Xiaohui Yan$^{2}$\thanks{Corresponding author.} \\
$^{1}$University of Chinese Academy of Sciences \\
$^{2}$Huawei Technologies Co., Ltd. \\
\texttt{qiulongpeng23@mails.ucas.ac.cn} \\
\texttt{\{liting142, maoshuai5, yangnan16, yanxiaohui2\}@huawei.com}
}

\begin{document}
\begin{CJK}{UTF8}{gbsn} 

\maketitle
\renewcommand{\abstractname}{Abstract}
\begin{abstract}
Input errors in question-answering (QA) systems often lead to incorrect responses. Large language models (LLMs) struggle with this task, frequently failing to interpret user intent (misinterpretation) or unnecessarily altering the original question's structure (over-correction).
We propose QuestionRAG, a framework that tackles these problems. To address misinterpretation, it enriches the input with external knowledge (e.g., search results, related entities). To prevent over-correction, it uses reinforcement learning (RL) to align the model's objective with precise correction, not just paraphrasing.
Our results demonstrate that knowledge augmentation is critical for understanding faulty questions. Furthermore, RL-based alignment proves significantly more effective than traditional supervised fine-tuning (SFT), boosting the model's ability to follow instructions and generalize. By integrating these two strategies, QuestionRAG unlocks the full potential of LLMs for the question correction task.
\end{abstract}

\section{Introduction}

The performance and reliability of any Question Answering (QA) system are fundamentally constrained by a critical, universal challenge: errors inherent  within the input questions. This is not a minor issue but a central bottleneck that affects systems of all complexities. These pervasive errors originate from two inevitable sources. The first is system-induced, where automated processes like Automatic Speech Recognition (ASR) and Optical Character Recognition (OCR) misinterpret spoken or written words. The second is user-induced, which covers the vast and unpredictable spectrum of human expression, including simple typographical mistakes, the use of regional dialects or colloquialisms, and ambiguous phrasing.
An ill-posed question triggers a cascade of failures throughout the entire QA pipeline. It immediately compromises the initial stage of question comprehension, which in turn derails the subsequent analysis and information retrieval, ultimately leading to an incorrect or irrelevant answer. Therefore, the ability to proactively identify and correct these input errors is not just a desirable feature—it is a foundational requirement for building robust, effective, and trustworthy QA systems.

\begin{table}[t]
    \centering
    \scriptsize
    \begin{tabular}{Sc|Sc}
        \hline
        \textbf{Error type} & \textbf{question example}\\
        \hline  \hline
        phonetic error& \underline{高四}和李白有什么关系(高四\ensuremath{\rightarrow}高适) \\  
        graphemic  error& OpenAI \underline{01}模型(01\ensuremath{\rightarrow}O1) \\  
        missing word & \underline{界}m7和m9哪个好(界\ensuremath{\rightarrow}问界) \\  
        wrong order & 有哈弗车吗\underline{纯电版}（哈弗车有纯电版吗）  \\ 
        repeating word & \underline{哈哈马斯}的状况(哈哈马斯\ensuremath{\rightarrow}哈马斯)  \\  
        ill-expression & 手机\underline{SM}码(SM\ensuremath{\rightarrow}SN)  \\  \hline
    \end{tabular}
    \caption{Typical error types of the questions in Chinese QA systems. The underlined text is miss-spelling. The parenthetical text describes how to correct the error.}
    \label{tab:question-type}
\end{table}

The challenge of question correction is rooted in the diverse nature of errors, which can range from simple typographical mistakes to complex grammatical and semantic inaccuracies, as shown in Table \ref{tab:question-type}. Traditional methods, often built upon foundational models like BERT \cite{devlin2019bert} and sequence-to-sequence architectures \cite{sutskever2014sequencesequencelearningneural}, have attempted to address this by creating specialized solutions. These approaches, however, are heavily reliant on explicit rule sets, hand-crafted features, and bespoke model architectures tailored to specific error types, such as phonetic-based correctors or grammatical taggers \cite{hong-etal-2019-faspell, zhang-2021-Phonetic, omelianchuk-etal-2020-gector}. While effective for their intended narrow tasks, this specialization inherently limits their generalizability and scalability \cite{li2024-c-llm}. Consequently, there is a compelling need to develop a universal correction framework capable of addressing a broad spectrum of errors without requiring task-specific customization.

Large Language Models (LLMs) represent a promising paradigm for question error correction, capable of addressing diverse error types within a single, unified framework. Building on this promise, a line of recent work has emerged to apply LLMs to this challenge. \cite{li2023ineffectivenesslargelanguagemodels} first investigates the effectiveness of GPT-3.5 on Chinese Text Correction, and reveals that LLMs handle fluency well and are more fault-tolerant to input data quality, but still have a noticeable gap when compared to previous state-of-the-art, fine-tuned smaller models.
Their analysis result identified the main reasons:
\begin{itemize}
    \item While LLMs possess vast general knowledge, they often fail when it comes to highly specific or nuanced errors. This is particularly evident when correcting errors involving proper nouns or domain-specific terminology. As shown in Table \ref{tab:over-correction}, the LLM could not distinguish between "市大" (Shìdà - City University) and "师大" (Shīdà - Normal/Teacher's University). This correction requires specific background knowledge about place names that a general-purpose LLM may not have prioritized.  Moreover, as a text-only model, LLMs cannot access information about a character's pronunciation (e.g., Pinyin) or its visual structure, which is a major handicap for identifying and correcting these types of errors.  We collectively refer to the problem of a large model being unable to or incorrectly understanding a user's query due to insufficient background knowledge as \textit{misinterpretation}.
    \item LLMs are trained to generate the most probable or fluent text based on their massive training data. This causes them to "correct" expressions that are grammatically valid but less common, replacing them with more conventional phrases \cite{fang2023chatgpthighlyfluentgrammatical, wu2023chatgptgrammarlyevaluatingchatgpt}.
    As shown in Table \ref{tab:over-correction}, the LLM changed "remarried wife" (再婚妻子) to "ex-wife" (前妻).
    In the context of question correction, this behavior is a form of error. We refer this problems as \textit{over-correction}.
\end{itemize}

\begin{table}[t!]
    \centering
    \footnotesize
    \begin{tabular}{Sc| Sc|Sc}
        \hline
        \textbf{Error Type} & \textbf{example} & \textbf{question}\\
        \hline  \hline
       & input & 湖南\underline{市大}怎么走 \\ 
        misinterpretation & ground truth & 湖南\underline{师大}  怎么走 \\  
        & correction & 湖南\uwave{四大}怎么走 \\   \hline 
        & input &  \underline{摩克多}再婚妻子 \\
       over-correction & ground truth & \underline{默克多}再婚妻子 \\  
       & correction  &   \underline{默克多}\uwave{前妻} \\  \hline
    \end{tabular}
    \caption{Examples of the misinterpretation and over-correction problems in LLM-based question correction. In the first example, the LLM misinterpreted the term "湖南市大" as "湖南四大", but the "湖南四大" does not exit in the real word. Instead, the correct should be "湖南师大" , which is a normal university and pronounced exactly the same as the user's original input. In the second example, in addition to correcting the erroneous text ("摩克多" to "默克多"), the LLM also change the correct text from "再婚妻子" to "前妻", a needless modification that changed the meaning of the original question.}
    \label{tab:over-correction}
\end{table}

To address the  misinterpretation and over-correction issues of LLMs when applying to Chinese question correction, we propose a knowledge-augmented LLM approach, called QuestionRAG, which can handle various types of errors in a simple and unified way. Specifically, to overcome the misinterpretation problem, QuestionRAG leverages Retrieval-Augmented Generation (RAG)  \cite{lewis2021retrievalaugmentedgenerationknowledgeintensivenlp} by 
introducing rich external knowledge for each question, such as the search results, entity description, similar questions. 
As the retrieval process takes into account a combination of textual, visual, and phonetic similarities, QuestionRAG enables the LLM to generate a correction based solely on the retrieved context, eliminating the need for the LLM itself to possess visual or phonetic discrimination capabilities.

To further address the over-correction problems, we designed a reinforcement learning training method for the question error correction task. This method can automatically stimulate the LLM's intermediate thought processes for question correction, reducing the reliance on annotated samples for the reasoning process. 

Experimental results show that the model trained with RL significantly outperforms SFT-trained models, and the over-correction problem is substantially mitigated.

\section{Related Work}
Chinese text Correction (CTC) is a challenging task since thousands of characters that can have similar pronunciations or visual appearances. In question answer system, errors in user questions are highly varied—stemming from colloquial language, input devices, and environmental conditions—making the task significantly more challenging.

Most conventional methods are tailored to address specific error types with small models, such as BERT\cite{devlin2019bert} and Seq-to-Seq models\cite{sutskever2014sequencesequencelearningneural}.
For instance, SpellBERT \cite{ji2021spellbert} is a lightweight pre-trained model utilizing BERT with additional graph-based visual and phonetic features for Chinese Spelling Check. Similarly, other BERT-based approaches \cite{hong-etal-2019-faspell, zhang-etal-2020-spelling, cheng-etal-2020-spellgcn} have integrated phonological and morphological knowledge into contextual embeddings in BERT-like models for identifying and correcting textual errors. 
These task-specific models are limited by their parameter scale and capabilities, thereby underperform in low-frequency and complex semantic scenarios.

While recent Large Language Models (LLMs) show promise for this task, standard in-context learning approaches that directly prompt the model for corrections \cite{li2023ineffectivenesslargelanguagemodels, sun2024richsemanticknowledge} have performed poorly, even underperforming traditional small models \cite{li2023ineffectivenesslargelanguagemodels, zhou2024cllmlearncheck}. More effective strategies include task-specific pre-training \cite{zhou2024cllmlearncheck} and fine-tuning \cite{li2023ineffectivenesslargelanguagemodels}.
\cite{zhou2024simpleeffectivetrainingfreepromptfree} combines an LLM with a Minimal Distortion Model for scoring corrections, but this is limited to candidates of identical length to the error. Addressing this, \cite{liang2025mtcscretrievalaugmentediterativerefinement} proposed a multi-turn CTC framework to maintain length consistency. Nevertheless, all of these methods still suffer from problems of misinterpretation and over-correction.
 
Recently, few work has begun to explore the integration of RAG \cite{lewis2021retrievalaugmentedgenerationknowledgeintensivenlp} into LLM-based text correction. These approaches typically enhance model performance by retrieving relevant data, such as domain-specific entities \cite{pusateri2025retrieval} or labeled corrections from existing training sets \cite{liang2025mtcscretrievalaugmentediterativerefinement}. However, the applicability of them is often limited by their reliance on curated, domain-specific knowledge. In contrast, our method leverages broad, general knowledge drawn from public sources like the web and Wikipedia, supplemented by unlabeled questions.
Furthermore, to the best of our knowledge, our work is the first to integrate RAG with reinforcement learning (RL) to align an LLM specifically for the text correction task. This novel combination successfully alleviates the aforementioned problems of both misinterpretation and over-correction.

\section{Methodology}

We first introduce the QuestionRAG framework, and then discuss how reinforcement learning is utilized within it to address the challenge of over-correction.

\begin{figure*}[t!]
    \centering
    \includegraphics[width=16cm]{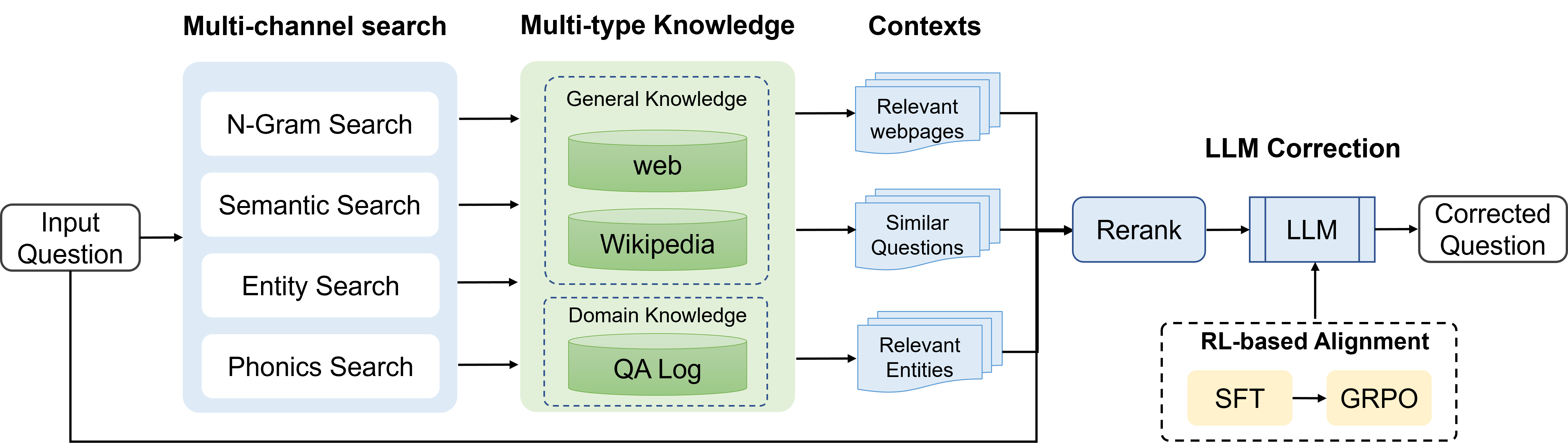} 
    \caption{The workflow of QuestionRAG includes a search stage  collecting relevant webpages, questions, and entities from external knowledge sources. It utilizes multi-facet search (n-gram for lexical similarity, embedding from semantic similarity, entity for conceptual similarity, and Pinyin for phonics similarity, etc.) from multiple knowledge sources (either general or domain-specific). With the search results as augmented knowledge, a LLM trained with reinforcement learning is utilized to generate the correction.}
    \label{fig:rag_flowchart}
\end{figure*}

\subsection{QuestionRAG}
Most prior research on LLM-based Chinese text correction has focused on enhancing the model's ability to understand and generate text based on character graphemes \cite{zhang2025unveilingimpactmultimodalfeatures} and phonetics \cite{yamashita2025llmbasedgenerativeerrorcorrection}. This task is inherently challenging, as large models are primarily trained for text generation. In contrast, the core idea of QuestionRAG is to fully leverage the LLM's strengths in text comprehension and generation, while offloading the processing of graphemes and phonetics—areas where LLMs are less proficient—to the retrieval stage. By integrating similarity factors such as string overlap, semantics, graphemes, and phonetics during search and ranking, QuestionRAG gathers relevant texts to serve as context, providing valuable guidance for the LLM to infer the correct correction.

We introduce general knowledge for a question by retrieval relevant texts from the following sources: 
\begin{itemize}
    \item \textbf{Web}: We take the question as a query, and obtain the search results form a commercial search engine. For simplification, we only kept the titles of relevant web pages as context.
    \item \textbf{Log of the QA system}: We search similar questions from historical questions extracted from the logs of a commercial question answer system. To filter low-quality questions, we filter ones with frequency less than 5.
    \item \textbf{Wikipedia}: we gather entities and their description from Wikipedia. To retrieval relevant entities, we first extract candidate entities from the input questions, and then search the entity base with these candidate entities.
\end{itemize}

QuestionRAG utilizes a multi-channel search system to handle multiple knowledge sources, as shown in Figure \ref{fig:rag_flowchart}. Its indexing stage employs a multi-faceted strategy combining three methods: (1) a ngram-based inverted index for lexical matching, (2) semantic vector similarity for conceptual retrieval, and (3) a Pinyin-based inverted index for phonetic matching. After an initial retrieval pass, the GTE-multilingual-reranker\cite{zhang2024mgte} is used to semantically filter and re-rank all candidate texts based on their relevance to the question. The resulting top-ranked texts then serve as the contextual input for the subsequent generation stage.

\subsection{Model Training with RL}

Following the paradigm of deepseek R1\cite{guo2025deepseek}, we adopt a two-stage training strategy consisting of Supervised Fine-Tuning (SFT) for cold-start and GRPO-based reinforcement learning for post-training.

\subsubsection{Cold-Start SFT}
To prepare our training data, we curated a small-scale, high-quality set by sampling both erroneous and error-free questions from QA logs. We manually corrected the erroneous samples to create their ground truth labels. Each question was then augmented with reasoning traces generated by deepseek R1\cite{guo2025deepseek}. During this process, any trace that led to an incorrect correction was discarded. Finally, to ensure the quality of the dataset, every training sample underwent a thorough manual review.

Instead of directly generating the corrected text, in cold-start SFT, the LLM first reason if the question contains an error, and then identifies the error's location if erroneous. Only after outputting this complete reasoning process does the model provide the final, corrected result. 

\subsubsection{GRPO Training}

To enhance the model's generalization and reasoning, we employ Group Relative Policy Optimization (GRPO) \cite{guo2025deepseek} in a post-training phase. This resource-efficient method eliminates the need for a separate evaluator model by directly comparing candidate responses, with each response scored by a predefined reward function. The design of this reward function is therefore critical for guiding the model toward high-quality, accurate corrections. Our reward function is built on two key components:
\begin{itemize}
    \item \textbf{Format Rewards.} This component systematically verifies whether model-generated outputs strictly adhere to predefined structural templates：$<$think$>$\{reasoning process\}$<$/think$>$\verb|\n\n|\{final answer\}. The reward function of output $\bm{c}$ is defined as:
\[
R_f(\bm{c}|\bm{q}) =
\begin{cases}
  1 & \text{if \textbf{c} format adhered} \\ 
  0 & \text{otherwise} \\
\end{cases}
\]

\item \textbf{Accuracy Rewards.} The accuracy reward is designed for a nuanced evaluation of the generated correction against the ground truth. Drawing inspiration from the use of Character Error Rate (CER) in error correction\cite{Leng2022SoftCorrectEC}, our function leverages a normalized edit distance\cite{Wang2024DANCERED} to quantify the similarity of output $\bm{c}$ to the target $\bm{g}$, defined as:
\[
\scriptstyle
R_a(\bm{c}, \bm{g}|\bm{q}) =
\begin{cases}
    2.0 & \text{if } \bm{c}=\bm{g} \\
    (1 - d_c) + \beta (1 - d_c)^2 & \text{if} d_c < d_q \\
    -\lambda \cdot d_c & \text{if} d_c > d_q \\
    0.0 & \text{otherwise}
\end{cases}
\]
\end{itemize}
where $d_q$ and $d_c$
   represent the normalized edit distances between the ground truth $\bm{g}$ and the input question $\bm{q}$, and the model's output $\bm{c}$, respectively. A smaller edit distance indicates a closer proximity to ground truth.
 $\beta$ is a hyperparameter that scales a non-linear bonus for improvements. $\lambda$ is a hyperparameter that scales the penalty for regressions.

The overall reward function summarizes the format reward and the accuracy reward, i.e.,
\[
R(\bm{c}, \bm{g}|\bm{q})=R_f(\bm{c}|\bm{q})+R_a(\bm{c}, \bm{g}|\bm{q}).
\]
The reward function described above encourages the gradual optimization of correction results during the RL training process, while penalizing incorrect corrections.
\section{Experiments}
In this section, we present the specific experiments conducted for error correction tasks.

\subsection{Datasets}
\paragraph{QCSet:} QCSet is a dataset derived from the logs of a commercial question-answering system. It comprises approximately 10,000 error-correction pairs, meticulously annotated by human experts to ensure high accuracy.
\paragraph{MCSCSet:}MCSCSet\cite{Jiang2022MCSCSetAS} is an open-source dataset focused on medical Chinese spelling correction. It is a large-scale dataset containing about 200k samples, manually annotated by medical specialists.
\paragraph{Qspell:}Qspell\cite{ye-etal-2025-qspell} is a public-domain dataset of approximately 250k error-correction pairs. It covers lots of topics, like formal terms, casual speech, and idioms, in both Chinese and English.

\subsection{Evaluation Metrics}
\textbf{Character Error Rate.}To quantitatively evaluate the performance of our error correction model, we employed the Character Error Rate (CER). CER is a widely used metric for assessing the accuracy of text transcription and correction tasks. It measures the minimum number of edits—such as insertions, deletions, or substitutions—required to align the generated output into the ground truth.

\subsection{Implementation Details}
We utilized Qwen3-8B\cite{qwen3} as our base model and trained it on the proprietary dataset QCSet. Specifically, for the cold-start phase, we  used around 1,000 QCSet samples enhanced with Chain-of-Thought reasoning generated via DeepSeek R1 distillation\cite{guo2025deepseek}. For the GRPO post-training phase, we employed an additional 7,000 QCSet samples, ensuring no overlap with the cold-start phase. This two-stage approach allowed us to progressively enhance the model's capabilities. 
To rigorously evaluate the model's performance, we tested it on three distinct test sets, each comprising around 1,000 randomly selected samples, with approximately 90\% of the samples requiring correction. These test sets were carefully chosen to ensure no overlap with the training data, providing a comprehensive assessment of the model's error correction capabilities.

In terms of training details, during the cold-start stage, we performed full parameter fine-tuning with a learning rate of 2e-6 and a total batch size of $8$. For the GRPO training stage, the learning rate is 5e-6 and batch size is 32. The hyperparameters in our reward function, $\beta$ and $\lambda$, were both set to 1.0 throughout our experiments. 

We also detail the computational requirements and training load for the post-training process, addressing the complexity of implementation. Our training, consisting of Cold-Start SFT and GRPO, was executed on Ascend 910B NPUs. The Cold-Start SFT stage involved training the model with 1k samples for 4 epochs on 8 NPUs, costing approximately 50 minutes. The subsequent GRPO phase, which represents the primary contributor to the total computational load, required training with 7k samples for 12 epochs on 32 NPUs, taking approximately 18 hours.

\begin{table*}[t!] 
    \centering
    \footnotesize
    \begin{tabular}{l|c|c|c} 
        \hline
        \textbf{Configuration} & \textbf{QCSet} &  \textbf{MCSCSet} & \textbf{QSpell}\\
        \hline \hline
        Original Question &16.13  & 16.35 & 14.78 \\
        No-RAG &15.37  &10.77 & 10.71 \\
        $\text{QuestionRAG}_{ICL}$(Entity) &14.31  &9.91 & 10.5\\ 
        $\text{QuestionRAG}_{ICL}$(Entity + Web Info) &13.6  &7.8 & 9.04 \\  
        $\text{QuestionRAG}_{ICL}$(Entity + Web Info + Similar Questions) & \textbf{12.53}  & \textbf{7.14} & \textbf{8.63}\\ %
        \hline
    \end{tabular}
    \caption{Comparison of $\text{QuestionRAG}_{ICL}$ with different types of knowledges(CER \%). No-RAG refers denotes Qwen3-8B without any knowledge argumentation. }
    \label{tab:rag_results}
\end{table*}

\renewcommand{\arraystretch}{1.2}
\begin{table*}[t!]
    \centering
\scriptsize
\begin{tabular}{c|c|c|c|c}
\hline
\textbf{Question} & \textbf{Similar Web page titles} & \textbf{Similar Questions} & \textbf{Related Entities} & \textbf{Correction} \\
\hline \hline
  湖南市大怎么走      & 湖南大学正门怎么走 & \mybold{到湖南师大怎么走}  & \mybold{湖南师大:湖南师范大学} & 湖南师大怎么走 \\
 &  湖南市政府怎么走  & \mybold{到湖南师范大学怎么走} &  &  \\
 \hline
 摩克多再婚妻子       &  \mybold{93岁默克多再婚} & 摩克多检测  &  \mybold{默克多:默克多也就是鲁伯特·默多克} &  默克多再婚妻 \\
                      &  摩克多塔拉  &  &  & \\
 \hline
\end{tabular}
    \caption{Two examples of question correction by QuestionRAG. Bold text represents the useful information that has been introduced. }
    \label{tab:case_study}
\end{table*}
\renewcommand{\arraystretch}{1}

\subsection{Results}

\subsubsection{In-Context Learning (ICL) Result}

To verify the effectiveness of knowledge argumentation of LLM on the question correction task, we evaluate QuestionRAG without any fine-tuning (referred to as $\text{QuestionRAG}_{ICL}$) by adding different types of knowledge into prompt, as detailed in Appendix \ref{sec:correction_prompt}. 
From Table \ref{tab:rag_results}, it is clear that progressively enhancing the RAG framework with additional knowledge sources consistently improved error correction performance on all the three dataset. 
Firstly, related entity information provided relevant or homophonic named entities and descriptions, which allowed the model to understand related concepts, especially by supplementing knowledge of rare entities. 
Secondly, the inclusion of web information provided real-time, diverse data, enhancing the model’s ability to handle novel errors. 
Finally, the introduction of similar questions from Our QA System also significantly improved error correction on the two open-source datasets. These findings suggest that external knowledge from either general or domain sources is crucial for LLM on the question correction task.

\subsubsection{Post Training Results}

\begin{table}[t!] 
    \centering
    \footnotesize
    \begin{tabular}{l|c|c|c} 
        \hline
        \textbf{Approach} & \textbf{QCSet} & \textbf{MCSCSet} & \textbf{QSpell}\\
        \hline \hline
        Original Question &16.13 & 16.35 & 14.78\\
        Fine-tuned T5  & 13.08  & 7.8 & 13.65\\
        $\text{QuestionRAG}_{ICL}$  & 12.53  & 7.14 & 8.63\\
        $\text{QuestionRAG}_{SFT}$  & 11.19 & 5.43 & 8.28\\
        $\text{QuestionRAG}_{GRPO}$  & \textbf{9.04} & \textbf{5.08} & \textbf{6.37}\\
        \hline
    \end{tabular}
    \caption{Comparing of SFT and GRPO against the other baselines on Chinese question correction(CER \%).}
    \label{tab:comparison_results}
\end{table}

We further aligned the LLM in QuestionRAG for question correction task with reinforcement learning. Its results are 
shown in Table \ref{tab:comparison_results}.
 $\text{QuestionRAG}_{GRPO}$ achieved the lowest CER on all the datasets, while 
  $\text{QuestionRAG}_{SFT}$ slightly outperforms  $\text{QuestionRAG}_{ICL}$. It suggests that both SFT and reinforcement learning alignment can further improve the effectiveness of QuestionRAG. Compared with SFT alone,
  reinforcement learning with GRPO exhibits stronger generalization capability. 
  Table \ref{tab:case_study} presents the relevant knowledge introduced for the incorrect questions, along with the useful information and the final corrected results. As can be seen, introducing additional context significantly reduces the difficulty for the LLM to perform the correction.

\renewcommand{\arraystretch}{1.2}
\begin{table*}[t!] 
    \centering
    \scriptsize
    \begin{tabular}{c|c|c|c}
        \hline
        \textbf{Question} & \textbf{Similar Web page titles} & \textbf{Similar Questions} & \textbf{Related Entities} \\
        \hline \hline
        终南山是谁 & 我如何来到终南山？& 终南山王维   & \mybold{钟南山}: 抗击非典、新冠疫情的领军人物\\ 
        \textit{ (Ground Truth: 钟南山是谁)}  & 谁都想到终南山隐居  & 西安终南山 & 终南山: 终南山是秦楚古道的所在地\\ 
       &   终南山的开派祖师是谁  &   终南山 &   终南山:《终南山》是王维创作的一首五律 \\
        \hline \hline
        \multicolumn{4}{l}{$\text{QuestionRAG}_{ICL}$: \textit{终南山是谁} (Incorrect); \quad 
            $\text{QuestionRAG}_{GRPO}$: \textbf{钟南山是谁} (\textit{Correct})
        } \\
        \hline
    \end{tabular}
    \caption{Robustness case study: $\text{QuestionRAG}_{GRPO}$ successfully corrects ambiguous queries despite highly noisy contextual information.}
    \label{tab:robustness_case_evidence}
\end{table*}
\renewcommand{\arraystretch}{1}

Beyond the quantitative improvements, we observed a surprising and critical qualitative advantage: $\text{QuestionRAG}_{GRPO}$ exhibits superior robustness even when faced with highly misleading and erroneous retrieval context. Table \ref{tab:robustness_case_evidence} clearly demonstrates this capability. The model successfully navigates a scenario where the majority of the entire retrieved context is erroneous and collectively favors common, yet wrong interpretations. The ability of $\text{QuestionRAG}_{GRPO}$to successfully filter this dominant, misleading information and adhere to the user's likely intent confirms that our approach guides the model toward a more reasoned and reliable correction trajectory.

\subsubsection{The impact of Model Scale}
\label{app: model_scale}

\begin{figure}[t!]
    \includegraphics[width=\columnwidth]{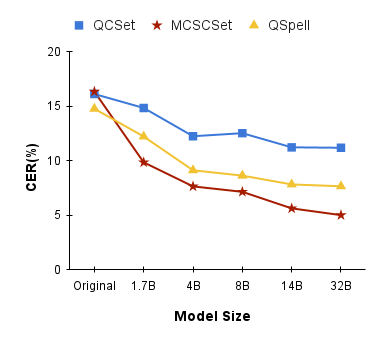} 
    \caption{Impact of model size on CER(\%) for Qwen3 models (1.7B to 32B) within the QuestionRAG framework, illustrating how increasing model size affects performance.}
    \label{fig:different_model_size_performance}
\end{figure}

We investigated the impact of model size within the QuestionRAG framework using various Qwen3 models. The results, presented in Figure \ref{fig:different_model_size_performance}, show that larger LLMs have a clear advantage over smaller ones, though this performance gap narrows as model size increases. This is because while all models share the same external knowledge provided by the RAG component, larger LLMs also possess more internal knowledge and stronger inherent capabilities for linguistic understanding and reasoning.

Moreover, a key finding demonstrates the power of knowledge augmentation: a smaller model equipped with QuestionRAG can outperform a much larger model that lacks it. For instance, a comparison between the "No RAG" baseline (Qwen3-8B) in Table  \ref{tab:rag_results} and the 1.7B parameter QuestionRAG model in Figure  \ref{fig:different_model_size_performance} reveals that the smaller model with QuestionRAG often achieves superior performance, highlighting that external knowledge can be more impactful than simply increasing model scale.


\section{Conclusion}

This paper introduces QuestionRAG, a novel framework that significantly enhances language model performance by addressing two of their core failures: misinterpretation and over-correction. Its primary innovation lies in a dual-pronged strategy: it leverages Retrieval-Augmented Generation to provide essential external context while employing Reinforcement Learning to meticulously align model behavior. This approach allows QuestionRAG to overcome the inherent deficiencies of LLMs in understanding character graphemes and phonetics, enabling it to fully exploit their strengths in text comprehension and generation within a single, unified framework that handles diverse error types. Beyond simple question correction, its methodology is directly applicable to broader tasks such as question rewriting, planning, and enhancing query understanding in search engines, opening new avenues for future research.

\section*{Limitations}
The performance of QuestionRAG is critically dependent on the quality of its search results. If the retrieved information is irrelevant or noisy, the accuracy of the final output will be compromised. Furthermore, because the RAG process significantly increases prompt length, it also increases latency, particularly the time to first token.

\bibliography{custom}

\clearpage
\appendix

\section{Prompt for Error Correction}
\label{sec:correction_prompt}


The complete version of the prompt used in our experiments is provided in the following. It outlines the instructions and principles for question error correction, including the use of retrieved information and strict adherence to correction rules.


\begin{tcolorbox}[colback=gray!5, colframe=gray!40,
  title=Prompt, fonttitle=\bfseries, breakable, enhanced,
  label={lst:correction_prompt},
  left=1mm, right=1mm, top=1mm, bottom=1mm
]
\ttfamily\footnotesize
You are a meticulous proofreading assistant. You need to combine your own knowledge with retrieved information (including similar questions, web titles, and entity information) to determine whether the user's original query requires correction. If no correction is needed, output the user's original query; otherwise, output the corrected query, strictly adhering to the following principles. Any changes must be based on phonetic similarity and preserve the framework of the user's original query.

\textbf{Correction Principles} (Strict Priority, Check from 1 to 4):

1. Minimal Modification Principle (Core): Only modify clearly erroneous parts of the user's original query (e.g., spelling errors, homophonic typos). Some special parts (including punctuation, spaces, capitalization, and word order) must remain unchanged. The corrected output should be similar in length and structure to the original query, with minimal changes. If the error cannot be identified, do not correct.

2. Corrections must strictly follow homophonic {or} near-homophonic rules (i.e., the corrected result must have the same or similar pronunciation as the original part). If an error is evident but does not satisfy the homophonic/near-homophonic rule, do not correct and output the user's original query.

3. Cautious Reference Principle (Supplementary Reference): Retrieved information (similar questions, web titles, entity information) is for reference only and may contain errors (especially similar questions, which may share the same mistakes). Use your own language knowledge to evaluate, and ignore retrieved information if it violates Principle 1 or 2 (e.g., suggests significant changes or alters pronunciation).

4. Semantic Preservation Principle (Non-Optimization Principle): Ensure the corrected query retains the same semantic meaning as the original, without altering the user's intent or structure for the sake of fluency. Only fix errors. If the intent is ambiguous, has multiple possibilities, or correction would distort the original meaning, do not correct (output the user's original query). Do not add or remove words to make the sentence more fluent.

\textbf{Output Requirement}: Directly output the corrected query or the user's original query without explanation. \\

\textbf{Retrieved Information:}

- Similar questions: 

\{similar questions\}
  
- Related Web Titles:

\{web titles\}
  
- Related Entity Information:

\{entity details\}

\textbf{User's Original Query:}

\{query\}

\textbf{Output the corrected result}
\end{tcolorbox}

\end{CJK}

\end{document}